\documentclass{article}

\usepackage[final,nonatbib]{neurips_2021}

\usepackage[utf8]{inputenc} 
\usepackage[T1]{fontenc}    
\usepackage{hyperref}       
\usepackage{url}            
\usepackage{booktabs}       
\usepackage{amsfonts}       
\usepackage{nicefrac}       
\usepackage{microtype}      
\usepackage{xcolor}         
\usepackage{graphicx}
\usepackage{multirow}

\title{A Data-Centric Approach for Training Deep Neural Networks with Less Data}
\author{%
  Mohammad Motamedi\\
  NVIDIA\\
  \texttt{mmotamedi@nvidia.com} \\
  \And
  Nikolay Sakharnykh\\
  NVIDIA\\
  \texttt{nsakharnykh@nvidia.com} \\
  \And
  Tim Kaldewey\\
  NVIDIA\\
  \texttt{tkaldewey@nvidia.com} \\
}

\begin{document}

\maketitle

\begin{abstract}
While the availability of large datasets is perceived to be a key requirement for training deep neural networks, it is possible to train such models with relatively little data. However, compensating for the absence of large datasets demands a series of actions to enhance the quality of the existing samples and to generate new ones. This paper summarizes our winning submission to the ``Data-Centric AI'' competition. We discuss some of the challenges that arise while training with a small dataset, offer a principled approach for systematic data quality enhancement, and propose a GAN-based solution for synthesizing new data points. Our evaluations indicate that the dataset generated by the proposed pipeline offers 5\% accuracy improvement, while being significantly smaller than the baseline.
\end{abstract}

\section{Introduction}
The emerging field of Data-Centric Artificial Intelligence (AI) is expected to deliver a set of techniques for dataset optimization, thereby enabling deep neural networks to be effectively trained using smaller datasets. This is an invaluable step towards democratizing AI, and will ultimately enables a broader set of applications to benefit from it. For example, applications with constrained data collection budgets, and domains such as flaw detection in manufacturing processes~\cite{fortune} where the training samples are inherently scarce. 

Even in the absence of a data-centric mindset, it is generally fair to assume that significant efforts are made in preparing any machine learning dataset to ensure it includes high quality data points. Nevertheless, one can always expect to find invalid instances including those that are mislabeled, ambiguous, or completely irrelevant. When the number of such samples is small compared to the cardinality of the dataset, their presence has negligible impact on performance. The reasons are twofold: First, gradients with respect to the model parameters are averaged in every mini-batch. This diminishes the impact of an undesirable set of gradients that are computed for an invalid input. Second, the effect on the averaged gradients is further limited and governed by the learning rate. As a result, when the vast majority of data is valid, the adverse impact of invalid samples becomes insignificant. This desirable property will disappear as the ratio of the invalid data points increases; a scenario that is more common in small datasets. In such settings, taking additional steps for improving the quality of a dataset will have a considerable impact on accuracy. 

Inspired by the first ``Data-Centric AI'' competition that was conducted and organized by Andrew Ng~et~al.~\cite{ng}, we present a principled approach that systematically enhances the quality of a dataset and helps with identifying invalid data points. In addition, we propose a model that utilizes Generative Adversarial Networks (GANs)~\cite{goodfellow2014generative} for synthesizing new samples. While the first part of this effort focuses on optimizing the existing data, the second part aims to generate new data points, thereby increasing the diversity of the training samples. We evaluate classifying handwritten Roman numerals and use the Roman-MNIST dataset~\cite{DCAIC} as our baseline.
\section{Problem statement}
\label{prob_stat}
Let us assume that we have a fixed neural network architecture, $\psi$, an undisclosed test set, $T$, an initial validation set, $V$, and an initial train set, $R$. Let us also assume that the size of the train set plus that of the validation set is expected to be less than a given constant that is noted by~$N_{max}$ in Equation~(\ref{eq_size_constraint}). This resembles the constraints that exist in some real-world applications such as flaw detection in steel industries where obtaining large samples is hard or impractical.
\begin{equation}
\label{eq_size_constraint}
|V| + |R| < N_{max}
\end{equation}
The question is: Given the aforementioned constraints, how should $R$ and $V$ be altered so that training and validating $\psi$ on them results in a model that yields a superior performance compared to the baseline when tested on $T$?

This paper answers the aforementioned question in the context of summarizing our effort targeting the first ``Data-Centric AI'' competition~\cite{ng}. In this challenge, $\psi$ is a cut off ResNet50~\cite{he2016deep} architecture, $N_{max}$ is 10,000, $|T| = 2420$, $|R| = 2067$, $|V| = 813$, and the test set was undisclosed during the competition. The target task is classification of handwritten Roman numerals and the performance metric is accuracy.

\section{Dataset optimization pipeline}
\label{Dataset_Optimization_Pipeline}
We propose a data optimization pipeline that investigates a dataset from various aspects, identifies invalid or mislabeled instances, and makes suggestions for addressing them. 

\textit{Step 1 – Duplicate Detection and Elimination}: The first step of the pipeline investigates the dataset to discover identical samples that may or may not have the same name. A multi-stage hashing approach is utilized to ensure that such an investigation has minimal execution time, even if the dataset includes millions of records. The benefits of this step are twofold: redundancy elimination and ensuring that the train and validation sets have no overlap.  

\begin{figure}
	\centering
	\fbox{\includegraphics[width=0.85\textwidth]{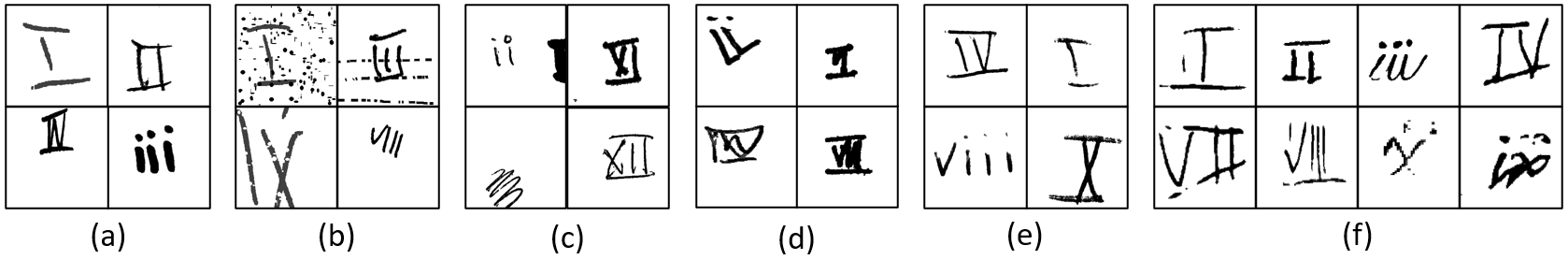}}
	\caption{(a) Samples from the initial dataset~\cite{ng}. (b) Samples of augmented images used for training the auxiliary models. (c) Samples of false positives instances detected by the auxiliary models. (d)~Samples with high loss values flagged for review by the auxiliary neural networks. (e) Samples of additional handwritten data gathered for this research. (f) Additional data generated by the GAN.}
	\label{sample_images}
\end{figure}

\textit{Step 2 – Training Auxiliary Models}: In this step, a small number of samples are randomly selected and inspected by a human supervisor to certify their validity and to ensure that they are labeled correctly. Subsequently, a classifier is trained using this data. Since the data size is small, applying a heavy data augmentation is essential. Appropriate augmentation techniques vary based on the data and the target application. In the case of Roman numerals, we used 180 samples for training and 20 samples for validation. For data augmentation, we used random rotation with a factor of 0.05, random contrast change with a factor of 0.5, random translation with a range of 20\% of image dimensions, random pen-stroke like black spots, random white spots, and finally random dashed lines with various lengths and orientations. Figure~\ref{sample_images}~(b) illustrates some of the augmented images. It is worth mentioning that models trained in this step are auxiliary neural networks that help with facilitating the dataset optimization. These models are no longer required after step 3.

\textit{Step 3 – Dataset Investigation}: The model prepared in the prior step is used for inference on the rest of the dataset and the samples are sorted based on their loss values. The first K samples that have the smallest loss values are perceived to be valid and to have correct labels based on the models' assessment. This perception is confirmed by a human supervisor. Subsequently, these K samples are added to the trainset and will be used in the next iteration. In addition to the first K samples, we also investigate the last L samples that are perceived to be invalid, have wrong labels, or are simply hard to classify. Analogous to the first step, a human supervisor checks the outputs, the new labels that are suggested by the model, and performs the necessary adjustments as needed. These samples, if not removed, will also be added to the trainset. Subsequently, a new model is trained as described in step two and the process reiterates between these two steps. In the case of Roman numerals, since the results are submitted to a competition, we repeated the process until all of the samples were certified by a human.

\textit{Step 4 – Class Imbalance Resolution}: In this step, additional data from each class is moved to a surplus dataset to ensure that all classes have the same number of samples. As we augment the data, these samples have the highest priority for moving back into the trainset.

\textit{Step 5 – N-fold Cross-Validation}: Subsequent to class imbalance resolution is the test set selection, which is performed in a random fashion. The rest of the data will be divided into N folds. For each fold, the trainset and the validation set are selected randomly. For the Roman numerals, N equals to 8.   

\section{GAN-based sample synthesis}
\label{gan_based_sample_generation}
\begin{figure}
	\centering
	\fbox{\includegraphics[width=0.85\textwidth]{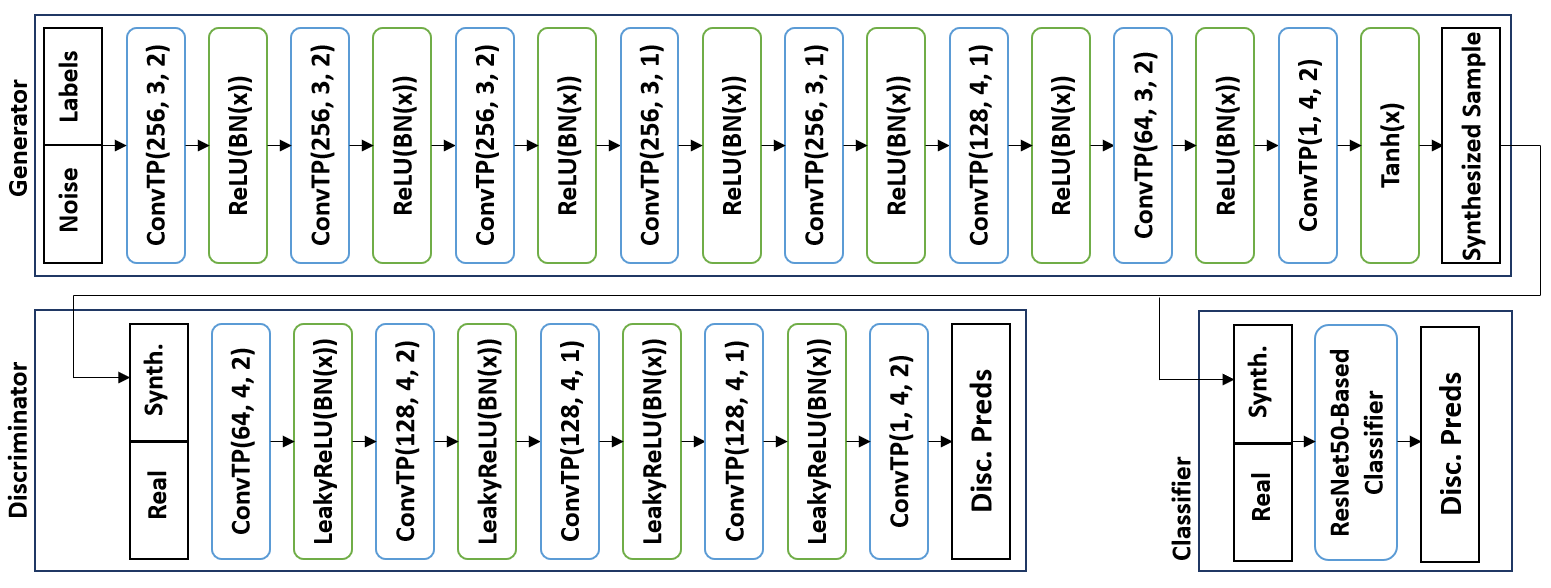}}
	\caption{Proposed GAN comprising of three components: A generator, a discriminator, and a pre-trained ResNet50-based classifier that is truncated from the ``conv2\_block3\_out'' layer.}
	\label{gan_arch}
\end{figure}
We designed a conditional GAN~\cite{mirza2014conditional} with an architecture analogous to that of the DCGAN~\cite{radford2015unsupervised} to generate additional samples. As illustrated in Figure~\ref{gan_arch}, the generator includes eight convolutional layers. All layers except the last one are followed by batch normalization and ReLU. The discriminator benefits from five convolutional layers and uses leaky ReLU with $\alpha=0.2$. The classifier is a cut off version of ResNet50~\cite{he2016deep} architecture that is truncated from the ``conv2\_block3\_out'' layer. The classifier loss as well as the discriminator loss are used to train the generator in a fashion that the synthesized samples are identified as real by the discriminator. However, they are misclassified by the classifier. As a result, the generated samples are expected to be edge cases where the classifier should benefit from more data. The loss function of the generator is shown in Equation~\ref{gen_loss}, where $\gamma$ and $\delta$ govern the impact of each component of the total loss. In this equation, $g$ denotes the ground truth and $y$ stands for the network output given the synthesized samples. The generator accepts a noise vector with the length of 64 that is sampled from a normal distribution with the mean of zero and standard deviation of one, and generates a grayscale output that is conditioned on the class labels. We use a batch size of 32, a learning rate of 0.0001, a pre-trained classifier, set $\delta = 1$, linearly increase $\gamma$ up to 0.5, and train the network for up to 200K iterations. As illustrated in Figure~\ref{sample_images}~(f), it \textit{appears} that the generated data are synthesized to be hard to classify. For example, the pen-stroke-like noises near ``I'' and ``VIII'' are the generator's attempts to make them look closer to ``II'' and ``IVIII'', respectively.
\begin{equation}
\label{gen_loss}
Gen~Loss = \delta \times BCE(g_{disc}, y_{disc}) - \gamma \times CCE(g_{class}, y_{class})
\end{equation}
\section{Experimental results}
\label{experimental_results}
\paragraph{Experimental setup}
All the reported times in this article are measured on a NVIDIA TITAN~V. Also, NVIDIA A100 GPUs are used for acceleration of GAN training. The final model and hyperparameters that are used for classifying Roman numerals are identical to the reference model of the first ``Data-Centric AI'' competition~\cite{ng}. The competition uses a ResNet50~\cite{he2016deep} architecture that is truncated from the ``conv2\_block3\_out'' layer. The batch size is 8, image size is 32x32, the network is trained for 100 epochs, and early stopping is applied for selecting the final model.  
\paragraph{Auxiliary models}
Similar to the final model, auxiliary ones use the reference architecture of the first ``Data-Centric AI'' competition. However, the training procedure of the auxiliary models differ from that of the competition in two aspects: First, auxiliary models are trained for 200 epochs. Second, augmentation is used as detailed in Section ~\ref{Dataset_Optimization_Pipeline}. We used K = 500 and L = 100 for processing the Roman MNIST~\cite{DCAIC} dataset and the results are summarized in Table~\ref{aux_perf_tbl}. In the first iteration, only 180 samples are used to train the network and the predictions are 92\% accurate. That is, 92\% of the samples that are flagged as valid or invalid have been confirmed by a human supervisor. Figure~\ref{sample_images}~(c) illustrates some invalid samples that are flagged as valid by the network. Two of the illustrated samples (i.e., ``ii'', and ``viii'') have validity to a certain degree. Nevertheless, they are determined to be ambiguous by a human supervisor. Selected samples from the first round are further used for training the second auxiliary model and the process continues as detailed in Section~\ref{Dataset_Optimization_Pipeline}.
\begin{table}
	\caption{Performance of the auxiliary models that are used in the pipeline.}
	\label{aux_perf_tbl}
	\centering
  \begin{tabular}{cllccccc}
  	\toprule
  	\multirow{2}{*}{\rotatebox[origin=c]{90}{Round}}             & \multicolumn{2}{c}{Dataset Size} & \multicolumn{2}{c}{Accuracy}  &Pipeline & Ratio of& \multirow{2}{*}{Epoch Time (s)} \\
  	\cmidrule(r){2-3}\cmidrule(r){4-5} & Train & Validation & Train & Validation &Accuracy   & Validated Data &\\ \midrule
  	1                                  & 180   & 20  & 94\% & 95\%  & 92\%       & 7\%   & 0.731 $\pm$ 0.013 \\
  	2                                  & 654   & 86  & 96\% & 97\%  & 89\%       & 28\%   & 2.970 $\pm$ 0.010 \\
  	3                                  & 1155  & 153 & 96\% & 95\%  & 92\%       & 47\%   & 5.220 $\pm$ 0.008\\
  	4                                  & 1651  & 216 & 97\% & 94\%  & 89\%       & 69\%   & 7.037 $\pm$ 0.013\\ 
  	5                                  & 2162  & 284 & 92\% & 95\%  & 80\%       & 90\%  & 9.210 $\pm$ 0.013\\ 
  	\bottomrule
  \end{tabular}	
\end{table}

\begin{table}
  \caption{Impact of dataset optimization on performance.}
  \label{exp_rslt_table}
  \centering
	\begin{tabular}{lcccccc}
		\toprule
		                                         & \multicolumn{3}{c}{Dataset Size} &  \multicolumn{3}{c}{Accuracy}   \\
		\cmidrule(r){2-4}\cmidrule(r){5-7}
	    Name                                     & Train & Validation & Test  &  Train   & Validation & Test    \\ \midrule
		Baseline                                 & 2067  & 813        & 2420  &  95\%    & 68\%       & 64\% \\
		Pipeline Output                          & 1370  & 500        & 2420  &  100\%   & 74\%       & 69\% \\
		Additional Sampled Gathered              & 8229  & 500        & 2420  &  100\%   & 90\%       & 83\% \\
		Synthesized Samples                      & 9455  & 500        & 2420  &  100\%   & 92\%       & 84\% \\ \bottomrule
	\end{tabular}
\end{table}
\paragraph{Performance assessment}
The results are summarized in Table~\ref{exp_rslt_table}, where the baseline refers to the initial dataset of the competition. This dataset is optimized by the proposed pipeline and offers 5\% accuracy improvement, while being 1.54x smaller. The third row of Table~\ref{exp_rslt_table} demonstrates the impact of adding additional samples that are gathered directly from new handwritten digits or from machine fonts. The handwritten digits are gathered from volunteers in an anonymous fashion. Finally, the GAN that is proposed in Section~\ref{gan_based_sample_generation} is used along with noise vectors that are drawn from a truncated normal distribution with a mean of 0, standard deviation of 1, and a truncation threshold of 0.7 to generate 1226 new samples, and the achieved result is tabulated in the last row.
\section{Conclusion}
We proposed a principled approach for data quality enhancement. Using the Roman-MNIST dataset~\cite{DCAIC} as the baseline, we demonstrated that the output dataset generated by our pipeline improves the accuracy by 5\%, while being significantly (1.54x) smaller than the baseline. We also designed a GAN that uses feedback from the classifier as well as the discriminator to synthesize new samples which are hard to classify. Using samples generated by this GAN further improved the accuracy by one percent.

\begin{ack}
We would like to thank Yasi Mahboub, Ehsan Gholami, Mahya Saffarpour, and Kourosh Vali for their help with gathering additional data. We also would like to thank Andrew Ng, Lynn He, Dillon Laird, and all other organizers from DeepLearning.AI and LANDING~AI for conducting the Data-Centric~AI competition.
\end{ack}
\medskip
{
\small
\bibliographystyle{unsrt}
\bibliography{neurips_2021}
}
\end{document}